\documentclass[letterpaper]{article} % DO NOT CHANGE THIS
\usepackage{aaai2026}  % DO NOT CHANGE THIS
\usepackage{times}  % DO NOT CHANGE THIS
\usepackage{helvet}  % DO NOT CHANGE THIS
\usepackage{courier}  % DO NOT CHANGE THIS
\usepackage[hyphens]{url}  % DO NOT CHANGE THIS
\usepackage{graphicx} % DO NOT CHANGE THIS
\usepackage{natbib}  % DO NOT CHANGE THIS AND DO NOT ADD ANY OPTIONS TO IT
\usepackage{caption} % DO NOT CHANGE THIS AND DO NOT ADD ANY OPTIONS TO IT
\usepackage{algorithm}
\usepackage{algpseudocode}
\usepackage{amsmath}
\usepackage{amssymb}
\usepackage{subfigure}
\usepackage{tabularx}
\usepackage{booktabs}
\usepackage{multirow}   % 用于 \multirow
\usepackage{array}  
\usepackage{newfloat}
\usepackage{listings}
\DeclareCaptionStyle{ruled}{labelfont=normalfont,labelsep=colon,strut=off} % DO NOT CHANGE THIS
\floatstyle{ruled}
\newfloat{listing}{tb}{lst}{}
\floatname{listing}{Listing}
\title{Spiking Heterogeneous Graph Attention Networks}
\author{
    Buqing Cao\textsuperscript{{\rm 1}},
    Qian Peng\textsuperscript{{\rm 2},\thanks{Corresponding authors.}},
    Xiang Xie\textsuperscript{\rm 3},
    Liang Chen\textsuperscript{\rm 4},
    Min Shi\textsuperscript{\rm 5},
    Jianxun Liu\textsuperscript{\rm 3}\\
}
\affiliations{
    \textsuperscript{\rm 1}School of Computer Science and Artificial Intelligence, Hunan University of Technology, Zhuzhou, China \\
    \textsuperscript{\rm 2}School of Computer Science and Engineering,  Central South
 University, Changsha, China \\
    \textsuperscript{\rm 3}School of Computer Science and Engineering, Hunan University of Science and Technology, Xiangtan, China  \\
    \textsuperscript{\rm 4}School of Data and Computer, Sun Yat-Sen University, Guangzhou, China  \\
    \textsuperscript{\rm 5}School of Computing \& Informatics, University of Louisiana at Lafayette, LA, USA  \\
    \{buqingcao, pengqian369, xiangxie001, ljx529\}@gmail.com, chenliang6@mail.sysu.edu.cn, min.shi@louisiana.edu
}

\usepackage{bibentry}
\begin{document}

\maketitle

\begin{abstract}
Real-world graphs or networks are usually heterogeneous, involving multiple types of nodes and relationships. Heterogeneous graph neural networks (HGNNs) can effectively handle these diverse nodes and edges, capturing heterogeneous information within the graph, thus exhibiting outstanding performance. However, most methods of HGNNs usually involve complex structural designs, leading to problems such as high memory usage, long inference time, and extensive consumption of computing resources. These limitations pose certain challenges for the practical application of HGNNs, especially for resource-constrained devices. To mitigate this issue, we propose the Spiking Heterogeneous Graph Attention Networks (SpikingHAN), which incorporates the brain-inspired and energy-saving properties of Spiking Neural Networks (SNNs) into heterogeneous graph learning to reduce the computing cost without compromising the performance. Specifically, SpikingHAN aggregates metapath-based neighbor information using a single-layer graph convolution with shared parameters. It then employs a semantic-level attention mechanism to capture the importance of different meta-paths and performs semantic aggregation. Finally, it encodes the heterogeneous information into a spike sequence through SNNs, simulating bioinformatic processing to derive a binarized 1-bit representation of the heterogeneous graph. Comprehensive experimental results from three real-world heterogeneous graph datasets show that SpikingHAN delivers competitive node classification performance. It achieves this with fewer parameters, quicker inference, reduced memory usage, and lower energy consumption. Code is available at https://github.com/QianPeng369/SpikingHAN. 
\end{abstract}

% Uncomment the following to link to your code, datasets, an extended version or similar.
% You must keep this block between (not within) the abstract and the main body of the paper.
% \begin{links}
%     \link{Code}{https://github.com/QianPeng369/SpikingHAN}
%    % \link{Datasets}{https://aaai.org/example/datasets}
%    % \link{Extended version}{https://aaai.org/example/extended-version}
% \end{links}

\section{Introduction}

\begin{figure}[t]
\centering
\includegraphics[width=\columnwidth]{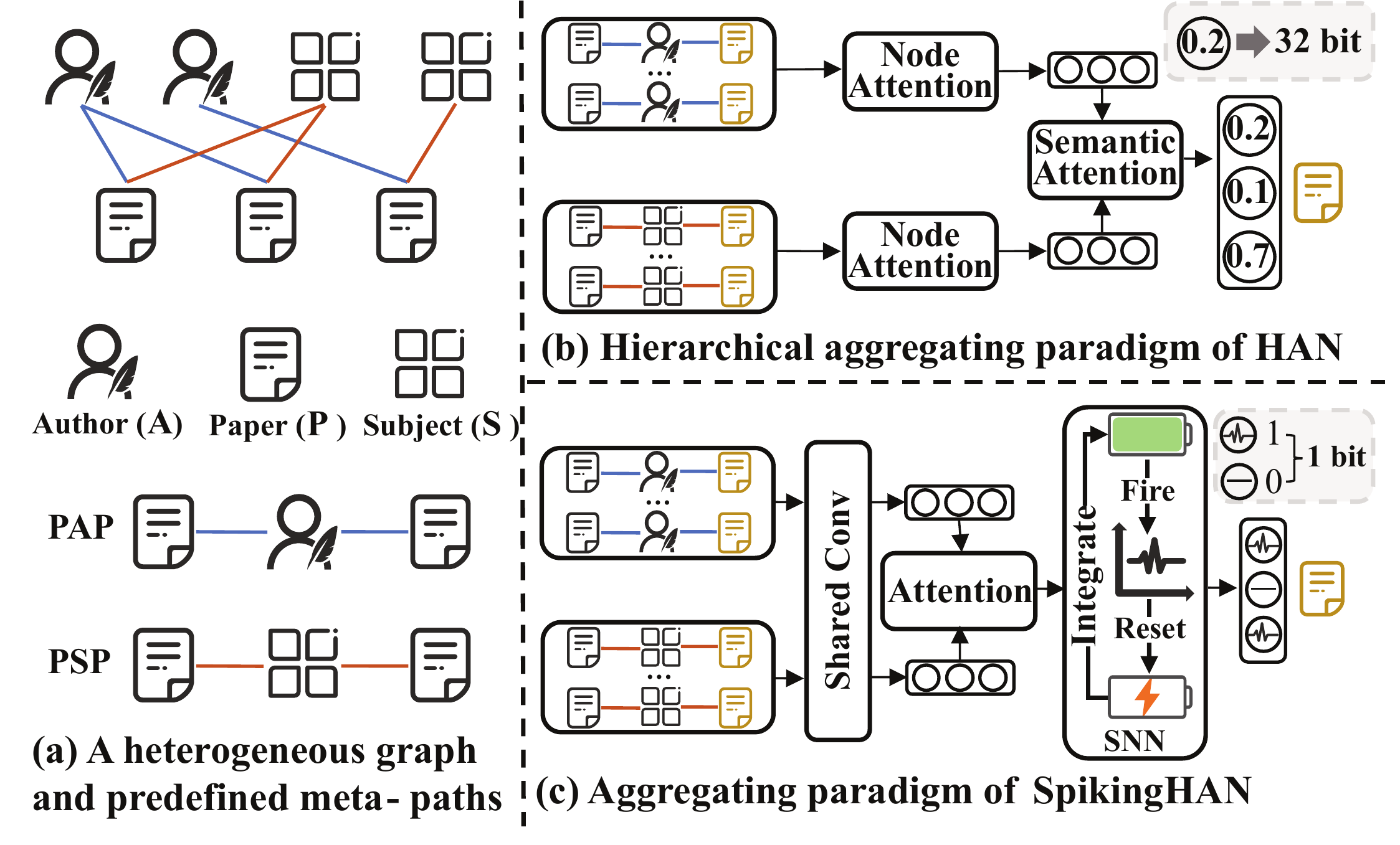}
\caption{Diagram of a heterogeneous graph and comparison between our model and HAN}
\label{fig1}
\end{figure}

Graph Neural Networks (GNNs) are excellent in combining graph data structures and node features, and have been widely used in various domains \cite{bib1,bib2,bib3,bib4}. Most GNNs are designed to learn node embedding vectors in homogeneous graphs, which consist of a single type of nodes and edges. However, real-world entity nodes and interaction edges often consist of multiple types, thereby forming heterogeneous graphs with rich structural and semantic information. Heterogeneous Graph Neural Networks (HGNNs) are specifically designed to handle heterogeneous graphs \cite{bib5}, leveraging meta-paths to capture the complex relationships among various types of nodes and edges. This not only improves the expression capability of the model but also enhances its adaptability to complex networks, showing great potential in fields such as social network analysis, recommendation systems, and bioinformatics \cite{bib6,bib7,bib8}. For instance, in the ACM paper dataset illustrated in Fig.~\ref{fig1}(a), the heterogeneous graph consists of three types of nodes and two types of edges. Using two predefined meta-paths, PAP and PSP, relationships between papers can be uncovered, such as papers sharing the same authors or papers belonging to the same topic.

Despite the excellent performance of HGNNs, many existing methods rely on complex structural designs, leading to problems challenges such as high memory usage, significant computational resource demands, lengthy inference time, and excessive power consumption during training. These limitations pose challenges for the deployment and expansion of HGNNs in practical applications, especially for resource-constrained devices \cite{bib9}. For example, Fig.~\ref{fig1}(b) illustrates the hierarchical aggregation paradigm employed by the widely-used HGNNs called HAN \cite{bib10}. HAN assigns a distinct node attention module for each meta-path, resulting in a significant increase in the model’s parameters, memory usage, and computational resource requirements as the number of meta-paths grows. Although this multi-level attention mechanism can effectively capture complex heterogeneous relationships, it intensifies the burden of computation and storage. Inspired by the brain's information processing methods, spiking neural networks (SNNs) use event-triggered and time-driven signals to update the parameters of neuron nodes, and can be categorized as brain-inspired networks characterized by discretization and sparsity \cite{bib11,bib12}. Different from traditional neural networks that pass messages through the neurons with floating-point value, the neurons of SNNs communicate in a sparse and binarized manner. These characteristics make SNNs highly suitable for application in low-energy consumption scenarios, mobile devices, and other resource-constrained environments.

Inspired by the successful application of SNNs in computer vision \cite{bib13}, researchers began to extend SNNs to graph data. Recent studies \cite{bib14,bib15,bib16,bib17} have integrated SNNs with graph convolutional networks, graph contrastive learning, and dynamic graphs, achieving competitive performance with reduced computational costs. These works also provide theoretical guarantees, demonstrating that binary spike-based graph networks possess expressive capabilities comparable to floating-point GNNs. This integration not only reduces energy consumption and computing resource usage, but also brings new possibilities for applying GNNs in resource-constrained environments. Although the application of SNNs to GNNs has gained increasing attention, their potential in heterogeneous graphs—a more prevalent graph data scenario in real-world applications—has yet to be fully explored and studied. Therefore, this paper fills this gap by designing effective SNNs for heterogeneous graphs.

This paper proposes a novel spiking heterogeneous graph attention network, called SpikingHAN, with its simplified aggregation paradigm illustrated in Fig.~\ref{fig1}(c). SpikingHAN introduces SNNs into the learning process of heterogeneous graphs and constructs binary heterogeneous graph representation in a compact and efficient manner. Unlike traditional HGNNs that require distinct aggregation mechanisms for each meta-path, SpikingHAN employs a single-layer graph convolution with shared parameters to perform aggregation for each meta-path, thereby significantly reducing the complexity of the model while maintaining the necessary expressive capability. SpikingHAN further uses the semantic-level attention mechanism to learn the importance weights of different meta-paths and performs targeted semantic aggregation based on these weights. Then, the Integrate, Fire, and Reset events of the SNNs are used to aggregate and update the spike signals at each time step. This process learns a sparse and effective binary representation, fully leveraging the advantages of SNNs for heterogeneous graph data and enabling fast inference. To conclude, the major contributions of this paper are summarized as follows:
\begin{itemize}
    \item A novel spiking heterogeneous graph attention network SpikingHAN is proposed, which innovatively combines SNNs and HGNNs. By simulating the spiking mechanism of biological neurons, it achieves low-energy and efficient computation, providing a new solution for the efficient processing of heterogeneous graphs.
    \item To our knowledge, SpikingHAN is the first attempt to integrate SNNs into heterogeneous graph data. Our work enables SNNs to be directly applied to heterogeneous graphs, especially in low-energy and resource-constrained environments, further promoting the development of heterogeneous graph-based applications.
    \item Comprehensive experiments on three real-world datasets indicate that, compared with similar models that output floating-point values, SpikingHAN achieves competitive node classification performance with fewer parameters, faster inference, smaller memory usage, and lower energy consumption.
\end{itemize}

\section{Related Work}

\subsection{Heterogeneous Graph Neural Network}
Existing HGNNs are generally categorized into metapath-based and relation-based models.
Metapath-based models maintain the heterogeneity of heterogeneous graphs by constructing neighbors using predefined meta-paths. MAGNN \cite{bib18} transforms node content features, incorporates intermediate nodes, and integrates multi-metapath semantics. THGNN \cite{bib19} extracts fine-grained topic-aware semantics by decomposing topics and leveraging global textual knowledge, improving link prediction and interpretability. HOAE \cite{bib20} applies a transformer-based mechanism to enhance attribute learning from heterogeneous neighbors and perform high-order attribute extraction via meta-paths. PHGT \cite{bib21} captures high-order semantics and long-range dependencies using node, semantic, and global tokens. In contrast, relation-based models avoid manual meta-paths by directly aggregating neighbor relation features. GTN \cite{bib22} adaptively learns edge types and composite relations without requiring domain knowledge. HetSANN \cite{bib23} processes multi-relation information via projection and attention, without predefined meta-paths. ie-HGCN \cite{bib24} automatically selects useful meta-paths through hierarchical aggregation, reducing preprocessing and computation. DAHGN\cite{bib25} addresses degree bias via dual-view contrastive learning between heterogeneous and homogeneous graph views. Although HGNNs perform well, most methods involve complex structural design, leading to high memory usage, long inference times, and large computing resource consumption. These limitations pose challenges for practical applications, especially on resource-constrained devices.

\subsection{Spiking Neural Networks}
Spiking Neural Networks (SNNs) bridge neuroscience and machine learning by simulating biological neurons with sparse, event-driven spikes \cite{bib26}. Due to their biologically plausible mechanisms and low energy consumption, SNNs are promising for energy-efficient applications. In computer vision, SNNs have shown strong potential. Spiking Transformer \cite{bib27} combines SNNs with self-attention, achieving state-of-the-art performance on ImageNet with improved energy efficiency. Spiking-YOLO \cite{bib28} introduces channel normalization and symbol neurons for accurate object detection in deep SNNs. \cite{bib13} redesigned semantic segmentation architectures like FCN and DeepLab using SNNs with surrogate gradient training, enhancing robustness and energy efficiency. Inspired by these successes, researchers began exploring SNNs on graph data. On homogeneous graphs, SpikingGCN \cite{bib15} integrates GCNs and SNNs, encoding graph structures into spike sequences and achieving excellent performance with energy efficiency on neuromorphic chips. GSAT \cite{bib29} generates sparse attention coefficients to enhance noise robustness in graph edge structures. SpikeGCL \cite{bib16} introduces SNNs into graph contrastive learning, producing efficient binarized 1-bit representations with low storage cost. On dynamic graphs, SpikeNet \cite{bib14} replaces traditional RNNs with spiking neurons to capture structural evolution efficiently. \cite{bib17} further improve dynamic graph representation by propagating early-layer information directly to the final layer and applying implicit differentiation to reduce memory usage. Although the application of SNNs to graph data has gradually attracted attention, SNNs have not been fully valued and studied in heterogeneous graphs, which are common in real-world scenarios. To address this, our proposed SpikingHAN applies SNNs to heterogeneous graphs and achieves quite competitive performance in node classification tasks with fewer parameters, faster inference, smaller memory usage, and lower energy consumption.

\section{Preliminary}
\subsubsection{Definition 1: Heterogeneous Graph.} A heterogeneous graph is defined as $\mathcal{G}=(\mathcal{V}, \mathcal{E})$, where $\mathcal{V}$ and $\mathcal{E}$ represent the sets of nodes and edges, respectively. The heterogeneous graph $\mathcal{G}$ is also associated with a node type mapping function $\phi:\ \mathcal{V}\rightarrow\mathcal{A}$ and an edge type mapping function $\psi: \mathcal{E} \rightarrow \mathcal{R}$. $\mathcal{A}$ and $\mathcal{R}$ represent predefined sets of node types and edge types, respectively, where $|\mathcal{A}|+|\mathcal{R}|>2$. 

\subsubsection{Definition 2: Metapath-based Neighbor.} A meta-path $\mathrm{\Phi}$ is a path of the form $A_1\overset{R_1}{\operatorname*{\to}}A_2\overset{R_2}{\operatorname*{\to}}\ldots\overset{R_l}{\operatorname*{\to}}A_{l+1}$, which describes the composite relationship $R=R_1\circ R_2\circ\ldots\circ R_l$ between the node types $A_1$ and $A_{l+1}$, where $\circ$ denotes the composite operator on this relationship. Given a meta-path $\mathrm{\Phi}$ in a heterogeneous graph, the metapath-based neighbors $N_i^\mathrm{\Phi}$ of node $i$ are the set of nodes connected to node $i$ through the meta-path $\mathrm{\Phi}$. If the meta-path $\mathrm{\Phi}$ is symmetric, then $N_i^\mathrm{\Phi}$ includes node $i$ itself.

\subsubsection{Definition 3: Spiking Neural Network.} SNNs usually have three core characteristics: (1) Integrate. Spiking neurons accumulate current through capacitors, gradually increasing the charge; (2) Fire. When the membrane potential reaches or exceeds the threshold $V_{th}$, the neuron generates a spike signal; and (3) Reset. After the spike signal is generated, the membrane potential is reset. There are generally two reset methods \cite{bib30}, one is to reset the membrane potential to a constant $V_{reset}$ (usually 0, and $V_{reset}<V_{th}$), and the other is to reset by subtracting the threshold $V_{th}$. The formal descriptions of these three core characteristics are respectively shown in equations (1), (2), and (3),

\begin{figure*}[t]
\centering
\includegraphics[width=.98\textwidth]{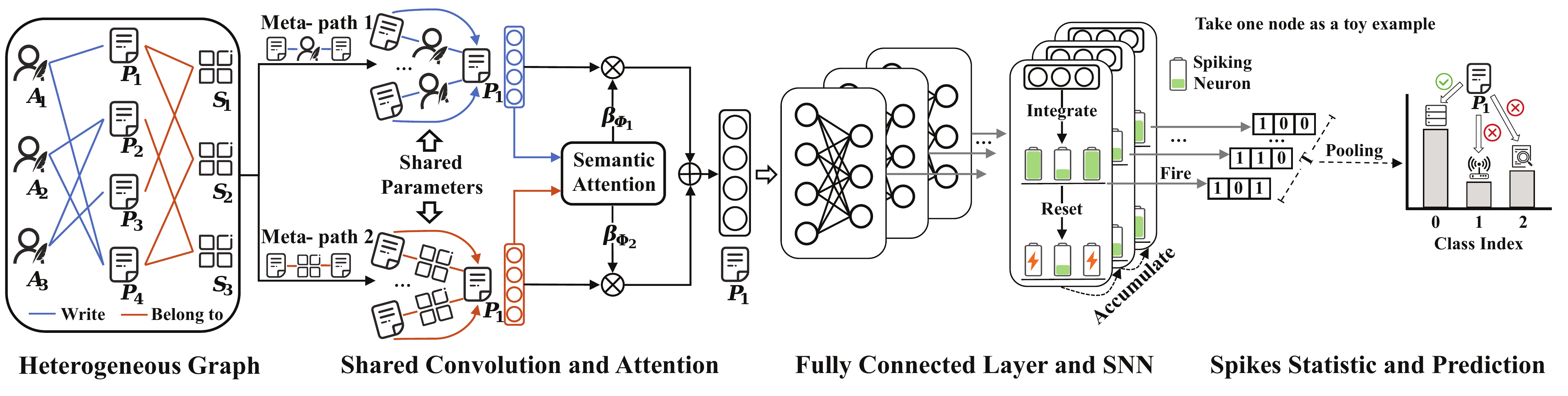}
\caption{The overall framework of SpikingHAN}
\label{fig2}
\end{figure*}

\begin{equation}
Integrate:V^t=\mathrm{\Psi}\left(V^{t-1},I^t\right)\label{eq1}
\end{equation}
\begin{equation}
Fire:S^t=\mathrm{\Theta}\left(V^t-V_{th}\right)\label{eq2}
\end{equation}
\begin{equation}\mathrm{Reset:}V^t=\left\{
\begin{array}
{c}S^t\cdot V_{reset}+(1-S^t)\cdot V^t \\
S^t\cdot(V^t-V_{th})+(1-S^t)\cdot V^t
\end{array}\right.\label{eq3}
\end{equation}
where $I^t$ and $V^t$ represent the input current and membrane potential at time step $t$, respectively. Whether a spiking neuron generates a spike signal is determined by the Heaviside function $\mathrm{\Theta}(\cdot)$, when $x\geq0$, $\mathrm{\Theta}(x)=1$, otherwise $\mathrm{\Theta}(x)=0$. The function $\mathrm{\Psi}(\cdot)$ is used to describe how spiking neurons receive input current and accumulate membrane potential. We can represent $\mathrm{\Psi}(\cdot)$ using the Integrate-and-Fire (IF) model \cite{bib31} and its variant the Leaky Integrate-and-Fire (LIF) model \cite{bib32}, as shown in equations (4) and (5), 
\begin{equation}
IF:V^t=V^{t-1}+I^t\label{eq4}
\end{equation}
\begin{equation}
LIF:V^t=V^{t-1}+\frac{1}{\tau_m}\left(I^t-\left(V^{t-1}-V_{th}\right)\right)\label{eq5}
\end{equation}
where $\tau_m$ is the membrane time constant used to control the rate of decay of the membrane potential, it usually needs to be adjusted manually. In the Parametric LIF (PLIF) model \cite{bib33}, $\tau_m$ can be automatically optimized during the training process to capture and learn different neuronal dynamics. This paper uses the surrogate gradient method to define $\Theta^{\prime}(x)\triangleq\sigma^{\prime}(\alpha x)$ during the loss backpropagation process \cite{bib14}, where $\sigma(\cdot)$ is the activation function, and $\alpha$ is the smoothing factor.

\section{Methodology}
This section introduces a novel heterogeneous graph neural network termed SpikingHAN. SpikingHAN tackles the challenge of semi-supervised node classification in a brain-inspired and energy-efficient manner. Its overall framework is shown in Fig.~\ref{fig2}, consisting of three main components, including the Shared Convolution and Attention, Fully Connected Layer and SNN, and Spikes Statistic and Prediction.

\subsection{Shared Convolution and Attention}
As one of the most representative models in GNNs, GCN \cite{bib34} is used in SpikingHAN to construct the shared graph convolution module that helps to aggregate each node's metapath-based neighbor information. Previous work usually assigns independent aggregation modules to each meta-path. Although this approach is effective, it may result in a significant increase in model complexity and is prone to overfitting problems when the number of convolution layers increases. In contrast, SpikingHAN uses shared parameters to perform metapath-based neighbors aggregation operations and executes only one layer of convolution operations, thus minimizing the model complexity while maintaining its expressive power. The single-layer graph convolution with shared parameters of SpikingHAN is formalized as shown in equation (6), 
\begin{equation}
h_i^{\Phi_p}=\sigma\left(\sum_{j\in N_i^{\Phi_p}}\frac{h_j^0\cdot W_1}{\sqrt{\widetilde{D}_i\cdot\widetilde{D}_j}}\right)\label{eq6}
\end{equation}
where $N_i^{\mathrm{\Phi}_p}$ is the set of neighbor nodes of node i based on the meta-path $\Phi_p\in\{\Phi_1,\Phi_2,\ldots,\Phi_P\}$. $\sigma(\cdot)$ is the activation function. $h_j^0$ is the initial feature representation of node $j$. $W_1\in\mathbb{R}^{d_{in}\times d_{hd}}$ is the learnable weight matrix, $d_{in}$ and $d_{hd}$ are the dimension of node initial feature and the hidden layer dimension respectively. $h_i^{\mathrm{\Phi}_p}$ is the new feature representation of node $i$ after the graph convolution operation with shared parameters. ${\widetilde{D}}_i$ and ${\widetilde{D}}_j$ are the degrees of node $i$ and node $j$ respectively. Given a set of meta-paths $\{\Phi_1,\Phi_2,\ldots,\Phi_P\}$, the initial node features will generate $P$ sets of node embedding $\{h^{\Phi_1},h^{\Phi_2},\ldots,h^{\Phi_P}\}$ with specific meta-path semantics after passing through the graph convolution with shared parameters.

After aggregating metapath-based neighbor information for each node, selectively integrating the semantics captured by different meta-paths is essential for learning more comprehensive node embeddings. Inspired by the semantic-level attention mechanism in \cite{bib10}, we adopt this attention mechanism to automatically determine the importance of various meta-paths and integrate them into their corresponding semantics. Specifically, the node embeddings of specific meta-path semantics are transformed nonlinearly, and their importance is measured by calculating the similarity between the transformed embeddings and the semantic-level attention vector $q$, formalized as shown in equation (7), 
\begin{equation}
I_{\mathrm{\Phi}_p}=\frac{1}{\left|\mathcal{V}\right|}\sum_{i\in\mathcal{V}}{q^T\cdot tanh\left(h_i^{\mathrm{\Phi}_p}\cdot W_2+b\right)}
\label{eq7}
\end{equation}
where $W_2\in\mathbb{R}^{d_{hd}\times d_{hd}}$ is the learnable weight matrix, $b$ is the bias vector. The importance weight $\beta_{\mathrm{\Phi}_p}$ of the meta-path $\mathrm{\Phi}_p$ can be obtained by applying Softmax normalization to the importance of all meta-paths, as shown in equation (8).
\begin{equation}
\beta_{\mathrm{\Phi}_p}=\frac{exp\left(I_{\mathrm{\Phi}_p}\right)}{\sum_{p=1}^{P}exp\left(I_{\mathrm{\Phi}_p}\right)}
\label{eq8}
\end{equation}

The learned importance weight $\beta_{\mathrm{\Phi}_p}$ can also serve as the attention coefficient for the meta-path to selectively integrate the various semantics revealed by the meta-path, as shown in equation (9),
\begin{equation}
H=\sum_{p=1}^{P}{\beta_{\mathrm{\Phi}_p}\cdot}h^{\mathrm{\Phi}_p}
\label{eq9}
\end{equation}
where $H$ represents the node embedding that integrates various meta-path semantics, and $P$ is the number of meta-paths.

\subsection{Fully Connected Layer and SNN}
The fully connected layer processes node embeddings $H$, which incorporate various meta-path semantics, and uses a trainable weight matrix to transform the high-dimensional feature space into a low-dimensional feature space. These low-dimensional features are subsequently used as input currents to the SNN. In the SNN, the input currents charge the spiking neurons, which then trigger a series of operations such as charge accumulation, spike fire and membrane potential reset.

Most deep SNN models use a multi-layer network structure that combines linear and nonlinear functions to process the input. However, based on the assumption of LightGCN \cite{bib35}, the depth of SNN is not a critical factor for predicting unknown labels on graphs. For this reason, we retain only the linear transformations in the fully connected layer and remove redundant modules such as nonlinear activation functions and biases to simplify the model and improve inference speed. The result $H\cdot W_3$ after linear transformation will be used as the input of SNN. First, the Integrate operation is performed, that is, charging the spiking neurons so that the charge gradually accumulates, as shown in equation (10), 
\begin{equation}
V^t=V^{t-1}+\frac{1}{\tau_m}\left(dropout\left(H\cdot W_3\right)-\left(V^{t-1}-V_{th}\right)\right)
\label{eq10}
\end{equation}
where $t=1,\ 2,\ ...,T, V^t$ is the membrane potential at time step $t$, $V^0=0$. $\tau_m$ is a learnable parameter used to control the decay rate of membrane potential. $W_3\in\mathbb{R}^{d_{hd}\times d_{out}}$ is the learnable weight matrix, $d_{out}$ is set to the number of classes of the target node.

When the accumulated membrane potential $V^t$ of the spiking neuron reaches or exceeds the given threshold $V_{th}$, the spiking neuron generates a spike through the Heaviside step function $\mathrm{\Theta}(\cdot)$, i.e., the Fire operation, as shown in equation (11).
\begin{equation}
\mathrm{\Theta}(V^t)=
\begin{cases}
1, & V^t\geq V_{th} \\
0, & V^t<V_{th} 
\end{cases}
\label{eq11}
\end{equation}

After a spiking neuron generates a spike, its membrane potential will release its potential like a biological neuron and begin accumulating voltage again, i.e., the Reset operation. In SpikingHAN, the membrane potential is reset by subtracting the threshold, as shown in equation (12).
\begin{equation}
V^t=\mathrm{\Theta}\left(V^t\right)\cdot\left(V^t-V_{th}\right)+\left(1-\mathrm{\Theta}\left(V^t\right)\right)\cdot V^t
\label{eq12}
\end{equation}

\subsection{Spikes Statistic and Prediction}
The spike sequence of each node at time step $t$ is generated by the spiking neurons. By applying average pooling on the outputs of the spiking neurons over multiple time steps, the class firing rate for each node can be obtained. This class firing rate will be used as the probability for the final node classification prediction, as shown in equation (13), 
\begin{equation}
\hat{y}=\frac{1}{\left|T\right|}\sum_{t=1}^{T}\mathrm{\Theta}\left(V^t\right)
\label{eq13}
\end{equation}
where $\hat{y}$ is the probability of the model classification prediction.

For the semi-supervised node classification task, this paper minimizes the cross-entropy loss through back propagation and gradient descent to optimize the weight parameters of the model. This cross-entropy loss of semi-supervised learning is shown in equation (14), 
\begin{equation}
\mathcal{L}=-\sum_{i\in\mathcal{V}_L} y_i\cdot ln\left({\hat{y}}_i\right)
\label{eq14}
\end{equation}
where $\mathcal{V}_L$ is the set of labeled node indices. $y_i$ is the one-hot encoding of the true label for node $i$. ${\hat{y}}_i$ is the probability of node $i$ classification prediction.

\section{Experiment}
To validate the effectiveness of SpikingHAN, we perform comprehensive experiments on three public real-world datasets, aiming to answer the following research questions: 
\begin{itemize}
    \item RQ1: How does SpikingHAN perform in node classification?
    \item RQ2: Does SpikingHAN have an advantage in terms of computational cost?
    \item RQ3: How do different configurations of spiking neurons and time steps affect the performance of SpikingHAN?
\end{itemize}

\subsection{Experimental Settings}
\subsubsection{Dataset description.}The experiments employ three commonly used heterogeneous graph datasets (i.e., DBLP, ACM, and IMDB) to evaluate the performance of SpikingHAN. More details of datasets are in Appendix B.

\subsubsection{Baselines.}SpikingHAN will be compared with three types of GNN models: (1) homogeneous GNNs, such as GAT \cite{bib36} and DAGNN \cite{bib37}. (2) homogeneous graph SNNs, such as SpikingGCN \cite{bib15} and SpikeGCL \cite{bib16}. and (3) heterogeneous GNNs, such as  HAN \cite{bib10} , HINormer \cite{bib38}, and PHGT \cite{bib21}. For the implementation details of our model and baselines, see Appendix C.

\begin{table*}[t]
\centering
\small
\renewcommand{\arraystretch}{1.3}
\setlength{\tabcolsep}{4pt}
\begin{tabular}{c|c|c|ccccccc|c}
\toprule
\textbf{Dataset} & \textbf{Metric} & \textbf{Tr. ratio} & \textbf{GAT} & \textbf{DAGNN} & \textbf{SpikingGCN} & \textbf{SpikeGCL} & \textbf{HAN} & \textbf{HINormer} & \textbf{PHGT} & \textbf{SpikingHAN} \\
\midrule
\multirow{6}{*}{\textbf{DBLP}} 
 & \multirow{3}{*}{Mi-F1} & 20\% & 91.4$\pm$0.4 & 91.9$\pm$0.7 & 88.7$\pm$0.6 & 90.6$\pm$0.2 & 93.1$\pm$0.1 & 93.6$\pm$0.1 & \textbf{93.7$\pm$0.2} & 93.7$\pm$0.1 \\
 &                        & 40\% & 91.9$\pm$0.3 & 91.6$\pm$0.9 & 90.4$\pm$0.3 & 91.3$\pm$0.3 & 93.6$\pm$0.1 & 94.2$\pm$0.2 & \textbf{94.6$\pm$0.2} & 93.8$\pm$0.2 \\
 &                        & 60\% & 92.9$\pm$0.4 & 91.8$\pm$0.9 & 90.3$\pm$0.3 & 91.5$\pm$0.3 & 93.7$\pm$0.2 & 93.7$\pm$0.3 & \textbf{94.3$\pm$0.2} & 93.7$\pm$0.1 \\
 \cmidrule{2-11}
 & \multirow{3}{*}{Ma-F1} & 20\% & 90.8$\pm$0.4 & 91.2$\pm$0.8 & 88.1$\pm$0.6 & 89.9$\pm$0.2 & 92.5$\pm$0.1 & 92.2$\pm$0.1 & \textbf{93.3$\pm$0.2} & 93.2$\pm$0.1 \\
 &                        & 40\% & 91.4$\pm$0.3 & 91.1$\pm$1.0 & 89.5$\pm$0.3 & 90.7$\pm$0.4 & 93.1$\pm$0.1 & 93.8$\pm$0.2 & \textbf{94.2$\pm$0.2} & 93.3$\pm$0.1 \\
 &                        & 60\% & 91.6$\pm$0.4 & 91.1$\pm$1.0 & 89.3$\pm$0.4 & 90.9$\pm$0.4 & 93.2$\pm$0.2 & 93.2$\pm$0.4 & \textbf{93.6$\pm$0.1} & 93.2$\pm$0.2 \\
\midrule
\multirow{6}{*}{\textbf{ACM}} 
 & \multirow{3}{*}{Mi-F1} & 20\% & 91.1$\pm$0.4 & 91.5$\pm$0.3 & 91.2$\pm$0.4 & 89.4$\pm$0.2 & 92.8$\pm$0.2 & 93.2$\pm$0.2 & \textbf{93.4$\pm$0.1} & 93.3$\pm$0.1 \\
 &                        & 40\% & 92.1$\pm$0.2 & 92.2$\pm$0.4 & 91.5$\pm$0.2 & 90.2$\pm$0.1 & 93.3$\pm$0.2 & \textbf{93.5$\pm$0.2} & 93.5$\pm$0.1 & 93.0$\pm$0.3 \\
 &                        & 60\% & 92.1$\pm$0.2 & 92.3$\pm$0.3 & 91.9$\pm$0.2 & 90.6$\pm$0.2 & 93.1$\pm$0.3 & 93.2$\pm$0.1 & \textbf{93.3$\pm$0.2} & 92.9$\pm$0.2 \\
 \cmidrule{2-11}
 & \multirow{3}{*}{Ma-F1} & 20\% & 91.2$\pm$0.4 & 91.2$\pm$0.4 & 91.2$\pm$0.5 & 89.4$\pm$0.2 & 92.9$\pm$0.2 & 93.1$\pm$0.2 & 93.1$\pm$0.1 & \textbf{93.4$\pm$0.1} \\
 &                        & 40\% & 92.1$\pm$0.2 & 92.0$\pm$0.4 & 91.4$\pm$0.1 & 90.2$\pm$0.1 & 93.3$\pm$0.2 & \textbf{93.4$\pm$0.2} & 93.4$\pm$0.1 & 93.1$\pm$0.3 \\
 &                        & 60\% & 92.1$\pm$0.2 & 92.2$\pm$0.3 & 91.8$\pm$0.2 & 90.5$\pm$0.2 & 93.1$\pm$0.3 & \textbf{93.2$\pm$0.2} & 93.1$\pm$0.2 & 92.9$\pm$0.2 \\
\midrule
\multirow{6}{*}{\textbf{IMDB}} 
 & \multirow{3}{*}{Mi-F1} & 20\% & 58.5$\pm$0.3 & 59.4$\pm$0.2 & 51.3$\pm$0.6 & 55.2$\pm$0.4 & 61.3$\pm$0.4 & \textbf{63.3$\pm$0.2} & 63.2$\pm$0.2 & 62.9$\pm$0.2 \\
 &                        & 40\% & 61.9$\pm$0.3 & 61.7$\pm$0.5 & 54.4$\pm$0.8 & 58.6$\pm$0.3 & 63.1$\pm$0.4 & 64.3$\pm$0.3 & \textbf{64.5$\pm$0.1} & 64.2$\pm$0.3 \\
 &                        & 60\% & 62.5$\pm$0.3 & 61.9$\pm$0.6 & 55.7$\pm$0.9 & 60.5$\pm$0.4 & 64.6$\pm$0.5 & 64.8$\pm$0.2 & 65.0$\pm$0.2 & \textbf{65.2$\pm$0.5} \\
 \cmidrule{2-11}
 & \multirow{3}{*}{Ma-F1} & 20\% & 58.3$\pm$0.4 & 59.1$\pm$0.2 & 50.4$\pm$0.7 & 54.7$\pm$0.4 & 61.1$\pm$0.4 & 62.9$\pm$0.2 & \textbf{63.0$\pm$0.2} & 62.7$\pm$0.1 \\
 &                        & 40\% & 61.7$\pm$0.4 & 61.3$\pm$0.6 & 54.1$\pm$0.9 & 58.2$\pm$0.3 & 62.8$\pm$0.3 & 63.8$\pm$0.2 & \textbf{64.1$\pm$0.1} & 63.7$\pm$0.4 \\
 &                        & 60\% & 62.1$\pm$0.4 & 61.2$\pm$0.6 & 55.3$\pm$0.9 & 60.2$\pm$0.4 & 64.3$\pm$0.5 & 64.4$\pm$0.2 & 64.6$\pm$0.2 & \textbf{65.1$\pm$0.4} \\
\bottomrule
\end{tabular}
\caption{Experimental results for node classification (\%). Results are the mean and standard deviation obtained by running 10 random seeds. The best result in each row is highlighted in bold. (Tr. Ratio: Training ratio, Mi-F1: Micro-F1, Ma-F1: Macro-F1)}
\label{tab2}
\end{table*}

\subsection{Performance Comparison (RQ1)}
Table~\ref{tab2} summarizes the classification results across different datasets and training ratios.  By analyzing the experimental results, we arrive at the following conclusions:

\begin{itemize}
    \item Heterogeneous GNNs typically outperform homogeneous GNNs and spiking GNNs on tasks involving complex structures and diverse node types, as they can model multi-type relationships and capture richer semantic information. In contrast, homogeneous models treat all nodes and edges uniformly, making it difficult to distinguish heterogeneous semantics and leading to critical information loss.
    \item The homogeneous graph SNNs (e.g., SpikingGCN and SpikeGCL) are much smaller than the traditional homogeneous GNNs (e.g., GAT and DAGNN) in terms of the number of model parameters, but there is no significant gap in performance. This indicates that introducing SNNs into GNNs can effectively reduce model complexity while maintaining competitive performance, further demonstrating the application value of spiking neural networks in graph neural networks.
    \item Combined with Table~\ref{tab3}, Fig.~\ref{fig3}, and Fig.~\ref{fig4}, it can be observed that traditional heterogeneous GNNs (e.g., HINormer and PHGT) achieve strong performance but incur high computational costs. In contrast, SpikingHAN significantly decreases the computing cost while still achieving or even partially exceeding the performance of these traditional methods. This advantage is due to the fact that SpikingHAN simulates the biological neuron spike firing mechanism to achieve low-energy and efficient computation.
\end{itemize}

\begin{table}[h]
\centering
\scriptsize  
\renewcommand{\arraystretch}{1.3}
\setlength{\tabcolsep}{3pt}  % 减少列间距
\begin{tabular}{c|cc|cc|cc}
\hline
\textbf{Method} & \multicolumn{2}{c|}{\textbf{DBLP}} & \multicolumn{2}{c|}{\textbf{ACM}} & \multicolumn{2}{c}{\textbf{IMDB}} \\
\cline{2-7}
 & Param & Memory & Param & Memory & Param & Memory \\
\hline
\textbf{GAT}        & 349,196   & 963.37    & 960,521   & 1994.92   & 1,572,873  & 1174.64   \\
\textbf{DAGNN}      & 43,400    & 917.80    & 239,878   & 1153.13   & 785,926    & 627.71    \\
\textbf{HAN}        & 292,868   & 1,676.8   & 1,985,283 & 542.39    & 6,419,715  & 450.17    \\
\textbf{HINormer}   & 7,348,964 & 6361.18   & 3,633,287 & 1093.61   & 8,577,159  & 2108.72   \\
\textbf{PHGT}       & 8,791,360 & 7156.53   & 5,579,065 & 2414.34   & 9,311,232  & 4459.98   \\
\hline
\textbf{SpikingHAN} & 15,201    & 45.2      & 128,385   & 137.32    & 102,593    & 220.19    \\
\hline
\end{tabular}
\caption{Number of model parameters and maximum GPU memory allocation (MB) during training }
\label{tab3}
\end{table}

\subsection{Runtime Complexity (RQ2)}
\begin{figure}[h]
\centering
\includegraphics[width=0.65\columnwidth]{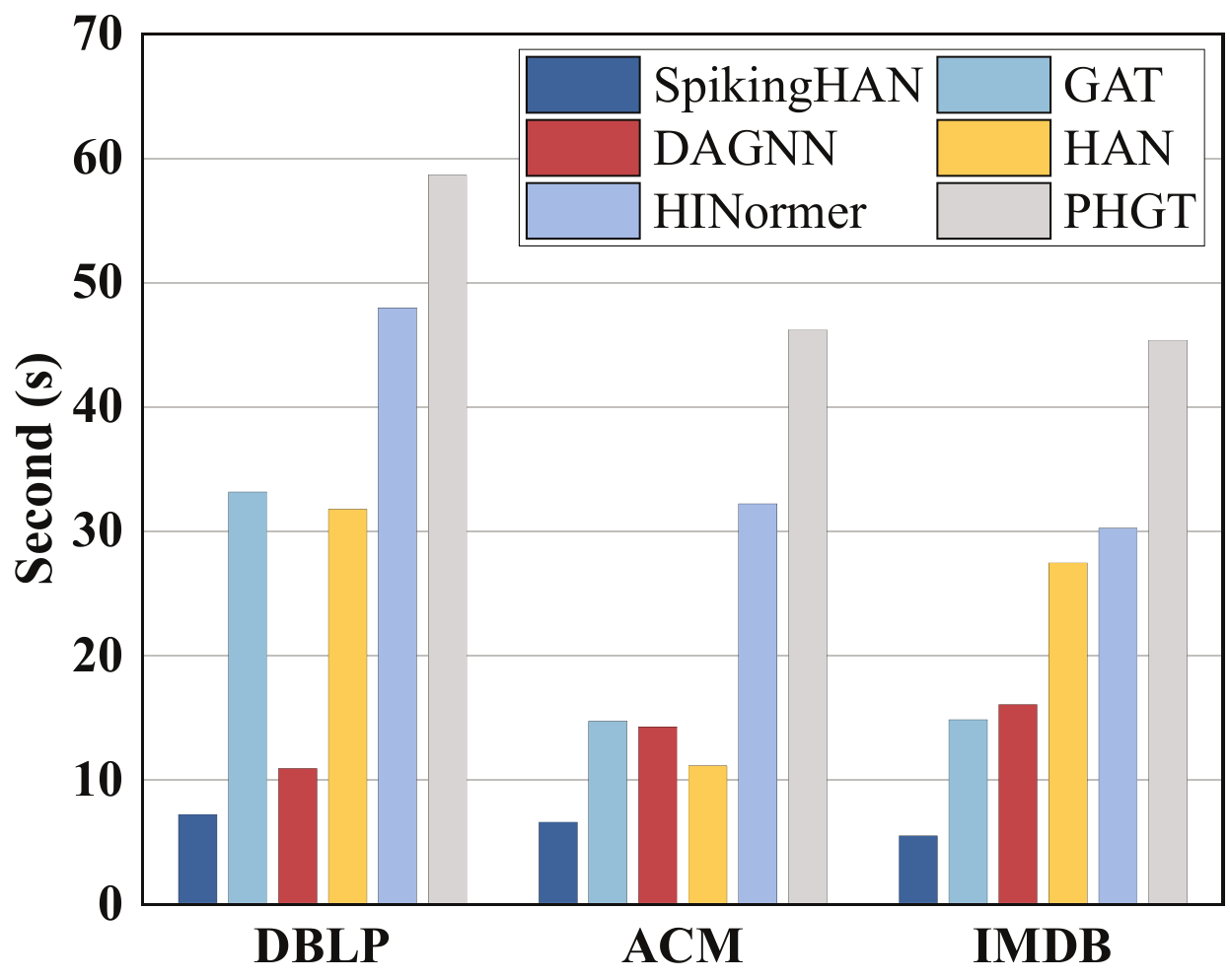}
\caption{Model training time on different datasets}
\label{fig3}
\end{figure}

\begin{figure*}[htbp]
\centering
\subfigure[DBLP]{
\begin{minipage}[t]{0.33\textwidth}
\centering
\includegraphics[width=\linewidth]{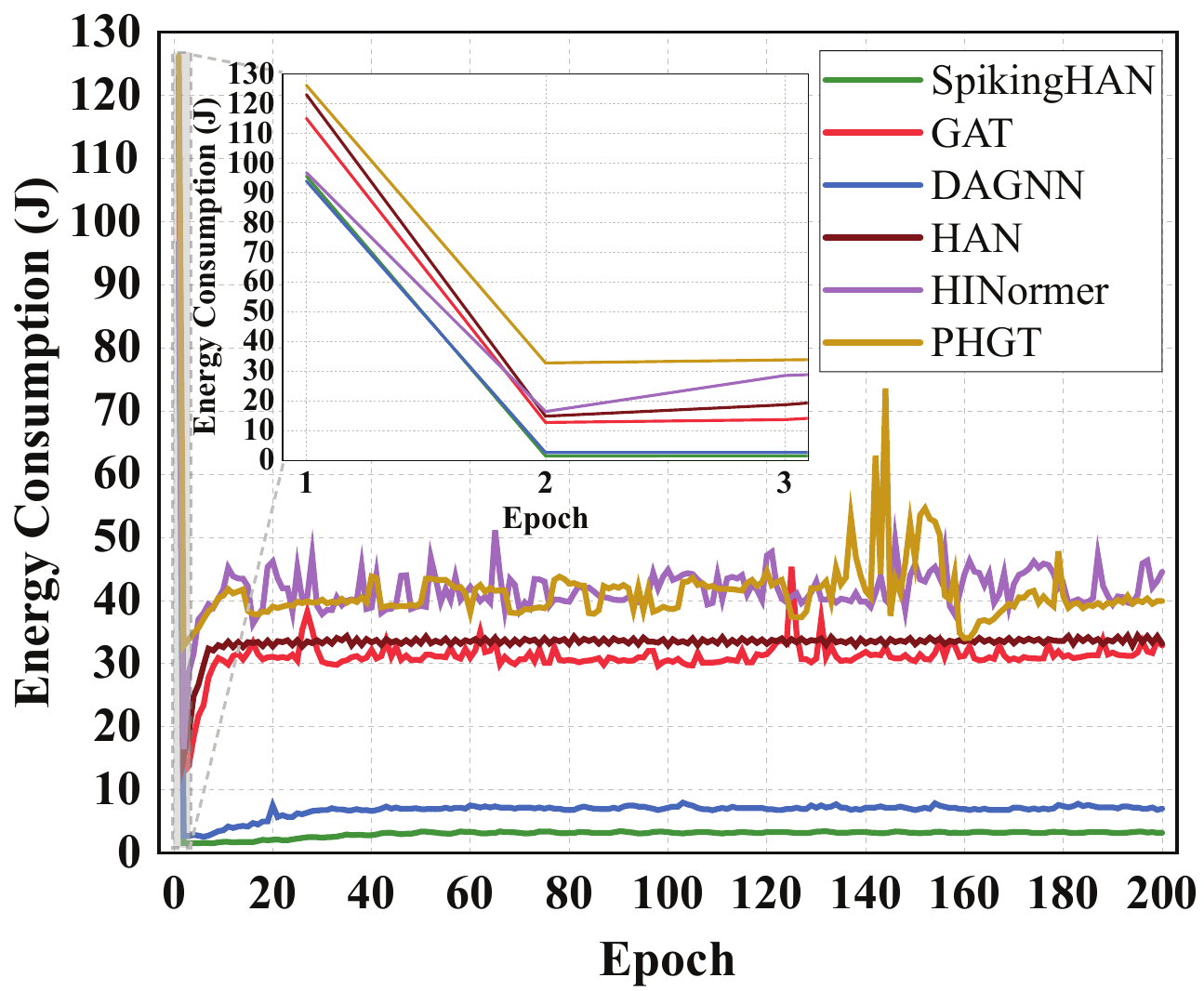}
\end{minipage}%
}%
\subfigure[ACM]{
\begin{minipage}[t]{0.33\textwidth}
\centering
\includegraphics[width=\linewidth]{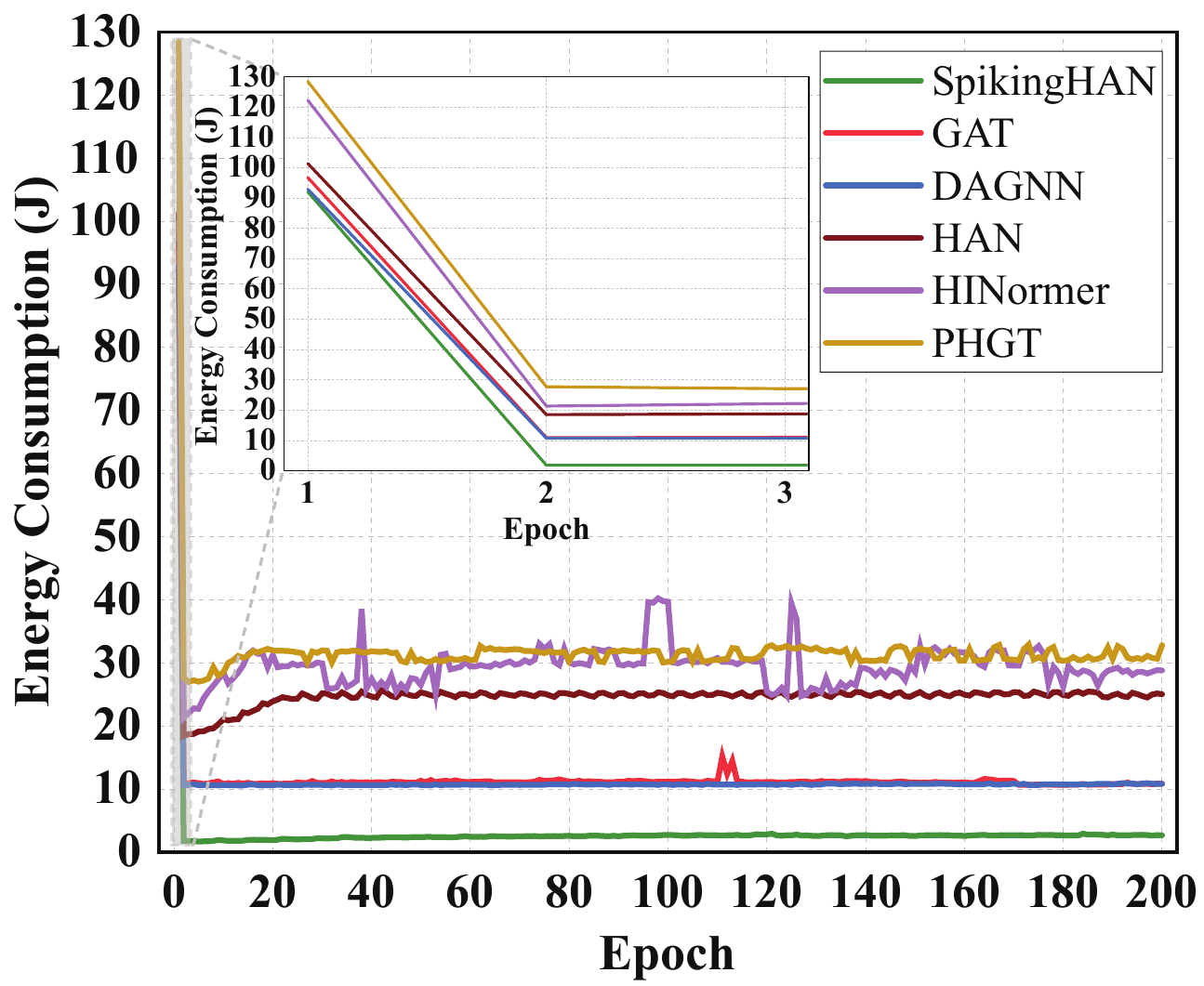}
\end{minipage}%
}%
\subfigure[IMDB]{
\begin{minipage}[t]{0.33\textwidth}
\centering
\includegraphics[width=\linewidth]{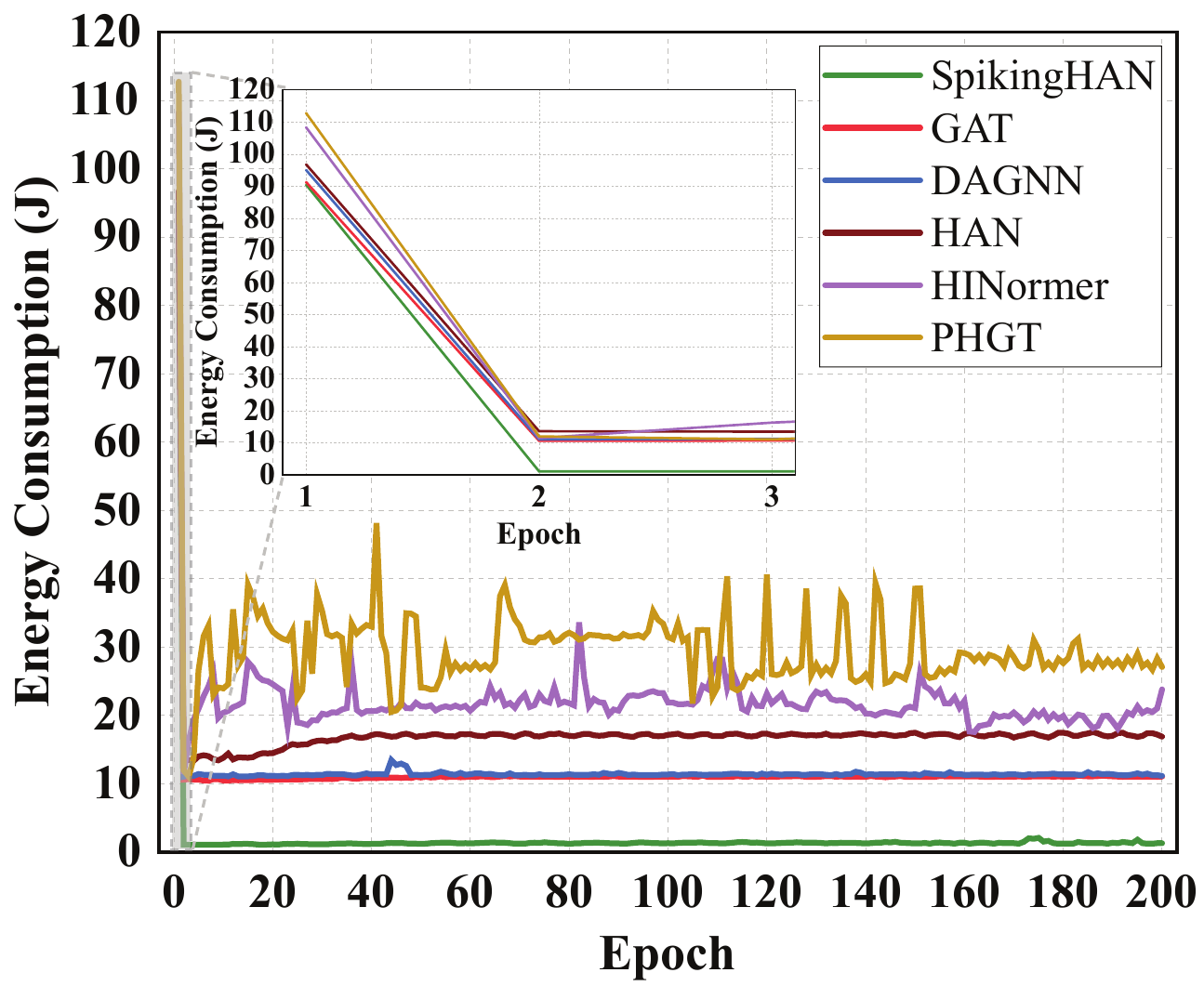}
\end{minipage}%
}%

\centering
\caption{GPU energy consumption per epoch during model training}
\label{fig4}
\end{figure*}

We further recorded the runtime complexity of SpikingHAN and partial baseline methods under different datasets, including the number of model parameters, the maximum GPU memory allocation during training, training time, and GPU energy consumption per epoch. These data are obtained through GPU monitoring and management library pynvml provided by NVIDIA, and are averaged based on the results of 10 different random seeds, as shown in Table~\ref{tab3}, Fig.~\ref{fig3}, and Fig.~\ref{fig4}. It can be seen from the data results that the number of parameters for SpikingHAN is significantly smaller than the other methods, while the number of parameters of heterogeneous graph neural network methods with excellent performance (e.g., HINormer and PHGT) is as high as millions. SpikingHAN achieves quite competitive performance with fewer parameters, which demonstrates that SpikingHAN is more efficient in terms of model complexity, leading to faster training speed and lower overfitting risk. At the same time, SpikingHAN has the lowest maximum GPU memory allocation on
all datasets, and this low memory requirement makes SpikingHAN more advantageous when memory resources are limited. In addition, Fig.~\ref{fig4} shows the GPU energy consumption per epoch during training on the three datasets. It can be observed that the energy consumption for the first epoch is higher compared to subsequent epochs. This is mainly because the first epoch usually involves model initialization and weight setting, while the subsequent epochs are mainly fine-tuning. Although PHGT performs best in classification effect, its GPU energy consumption is approximately 13 times that of SpikingHAN, and the performance improvement is not significant. The overall energy consumption of SpikingHAN on the three datasets is the lowest and relatively stable among all the compared methods. This demonstrates that SpikingHAN not only has excellent performance in dealing with heterogeneous graph learning tasks, but also can effectively reduce energy consumption, which has important economic and environmental value for large-scale applications and practical deployment.

\begin{figure}[h]
\centering
\begin{minipage}[t]{0.48\columnwidth}
\centering
\includegraphics[width=\linewidth]{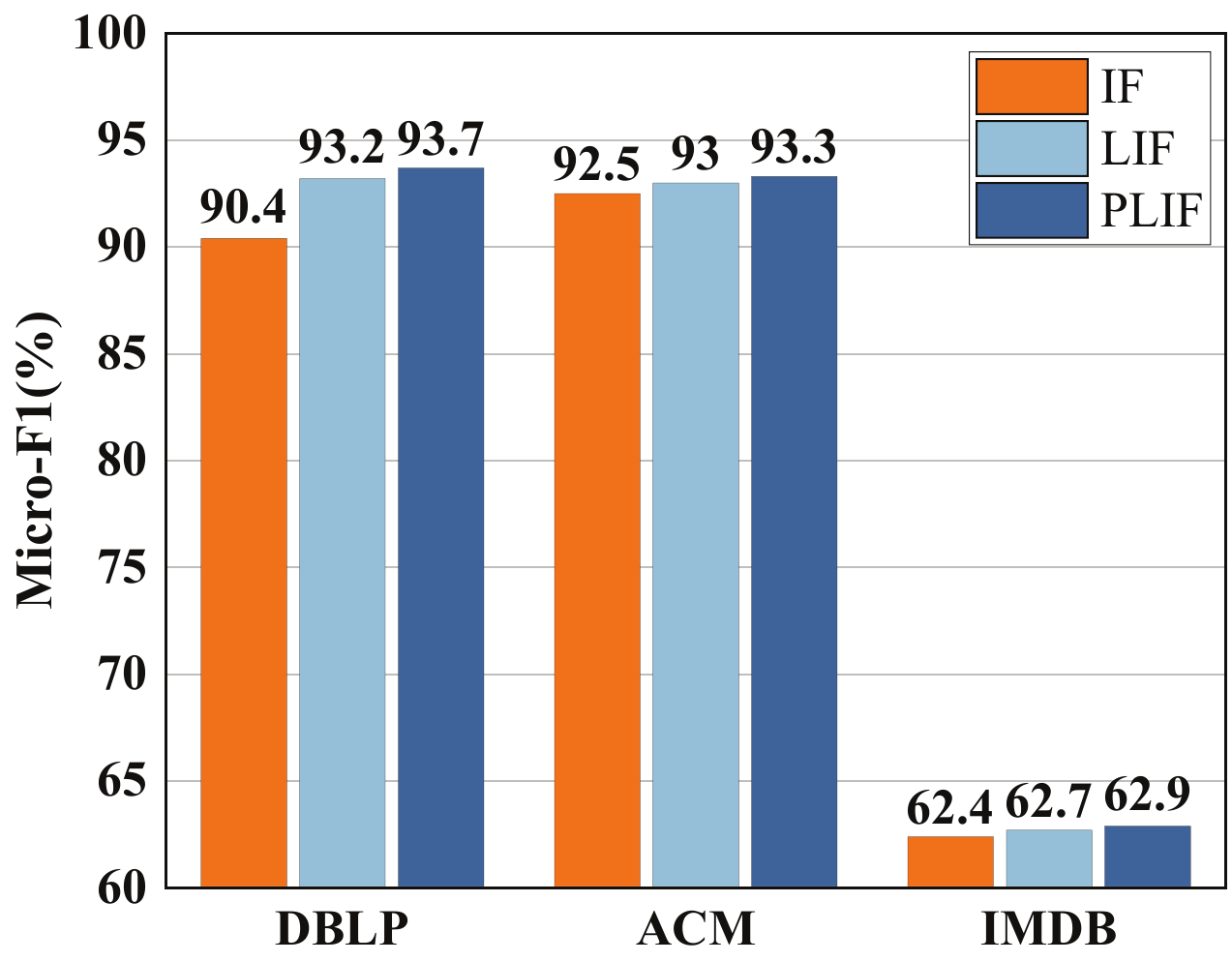}
\end{minipage}%
\hspace{0.02\columnwidth}
\begin{minipage}[t]{0.48\columnwidth}
\centering
\includegraphics[width=\linewidth]{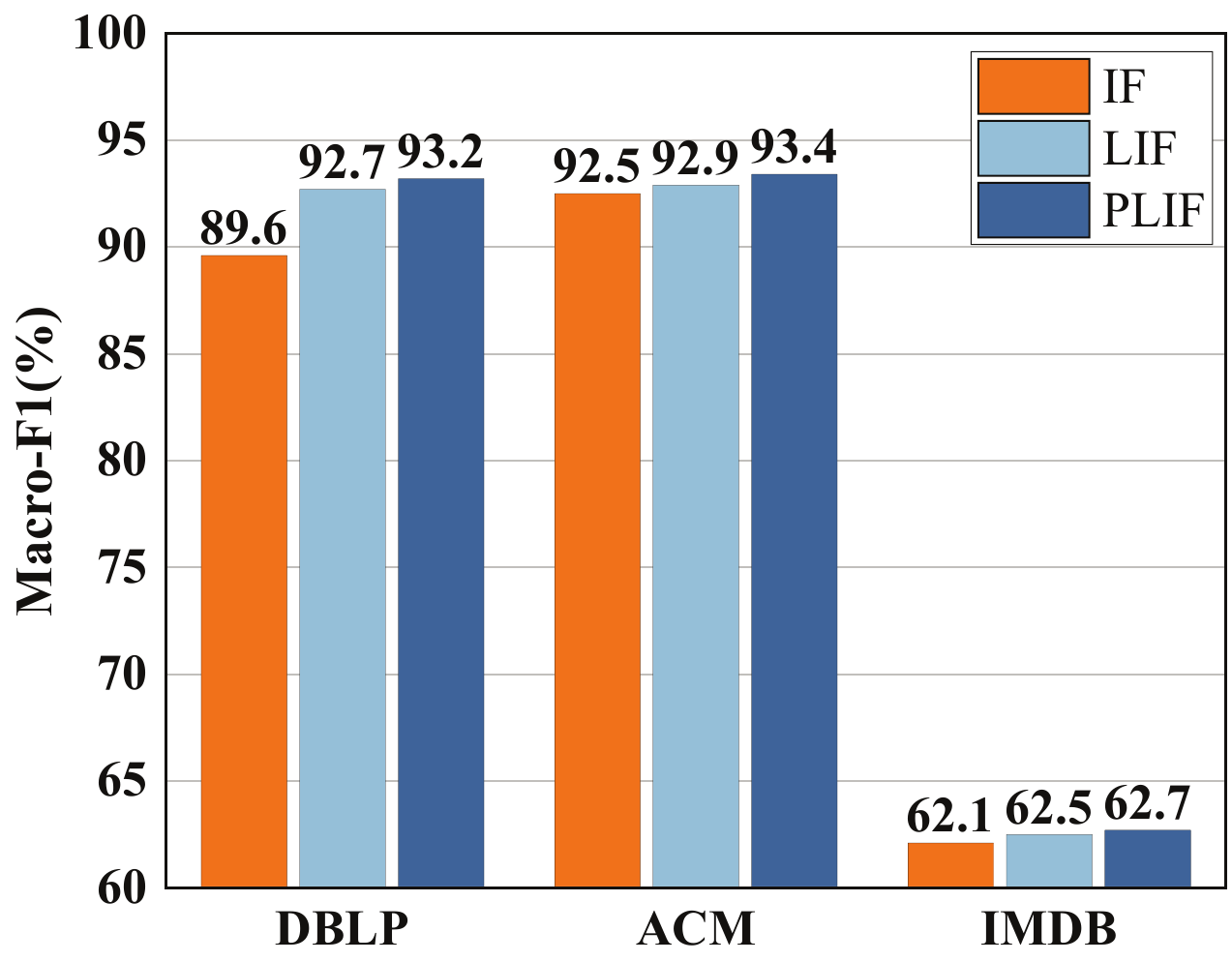}
\end{minipage}%
\caption{The impact of different spiking neurons}
\label{fig5}
\end{figure}

\begin{figure}[t]
\centering
\subfigure[DBLP]{
\begin{minipage}[t]{0.33\columnwidth}
\centering
\includegraphics[width=\linewidth]{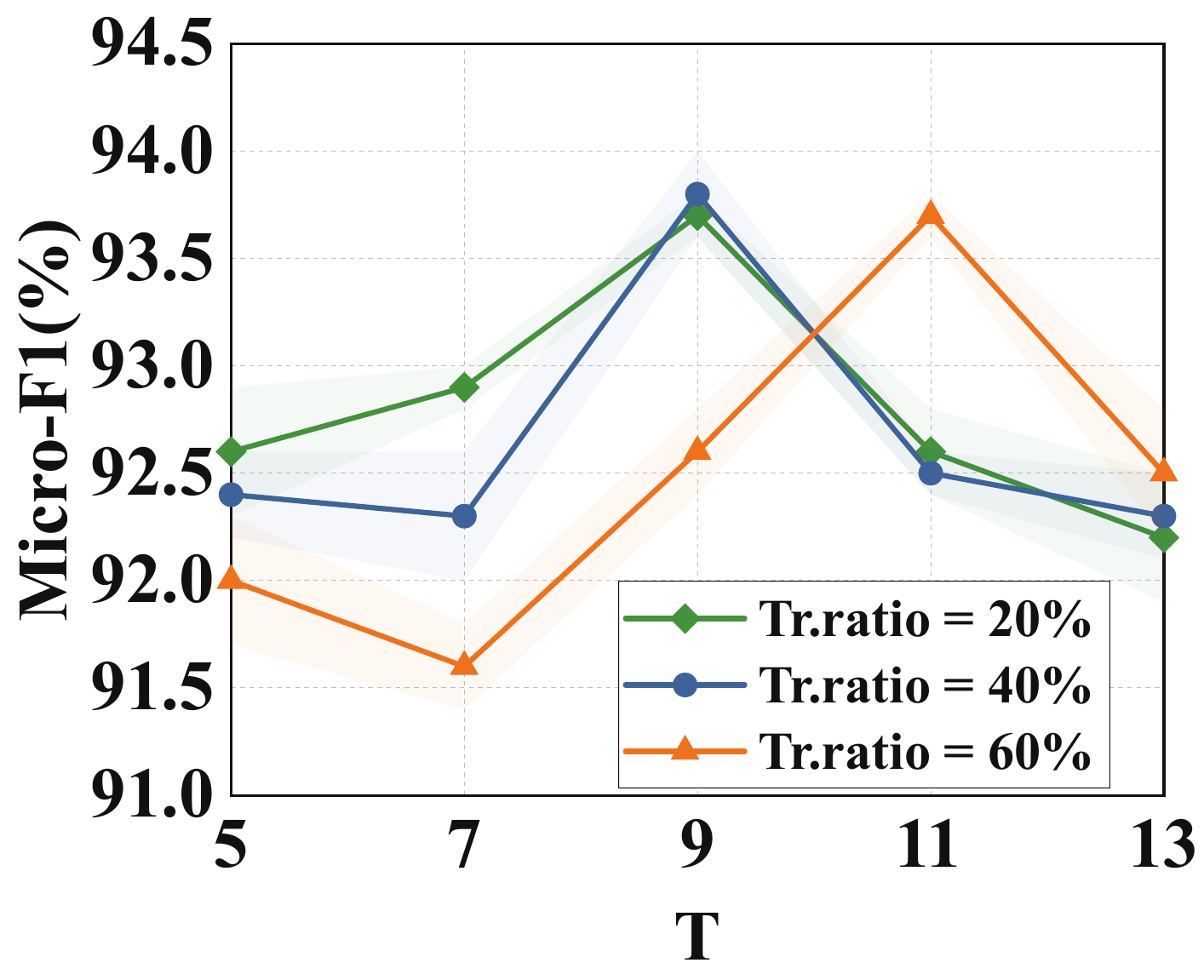}
\end{minipage}%
}%
\subfigure[ACM]{
\begin{minipage}[t]{0.33\columnwidth} 
\centering
\includegraphics[width=\linewidth]{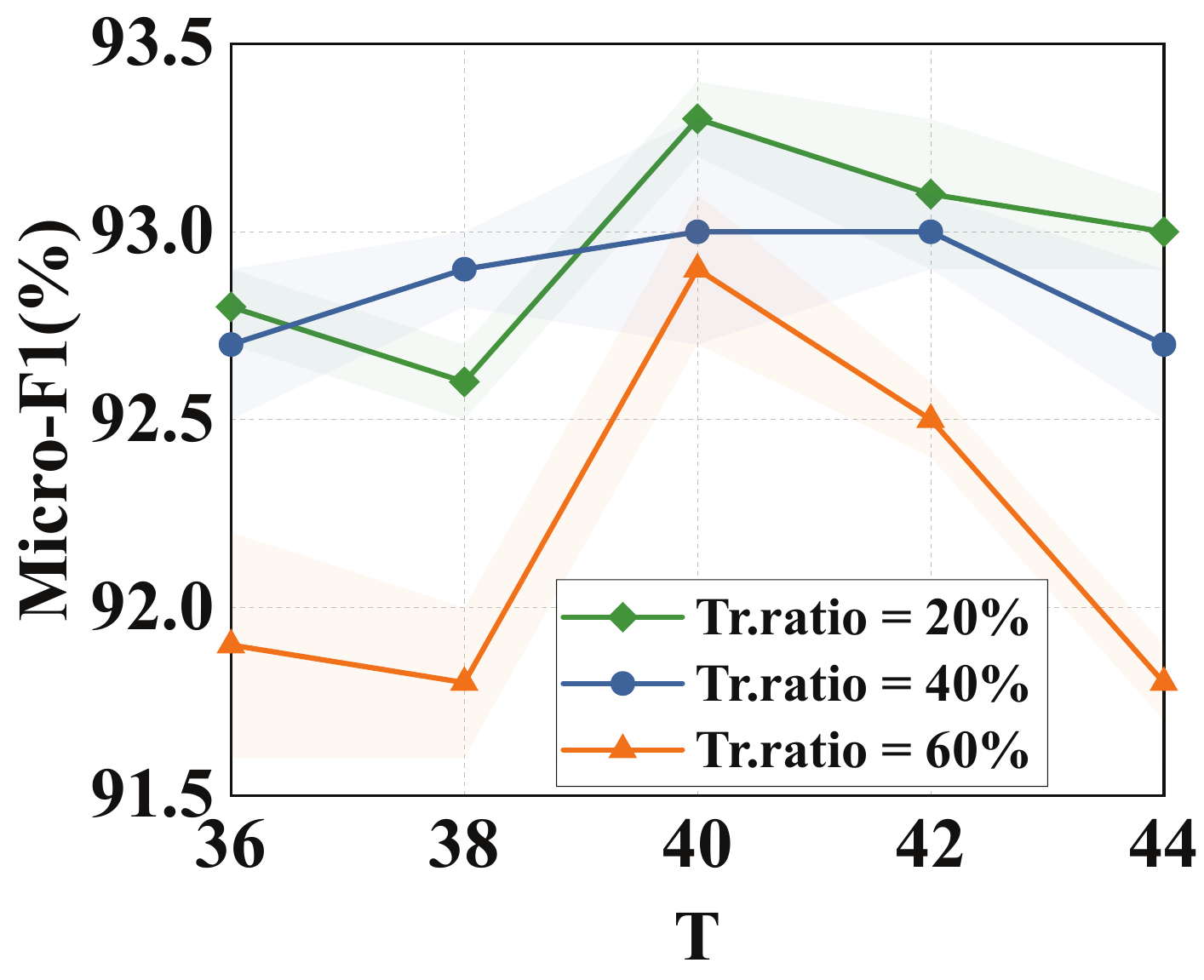}
\end{minipage}%
}%
\subfigure[IMDB]{
\begin{minipage}[t]{0.33\columnwidth}
\centering
\includegraphics[width=\linewidth]{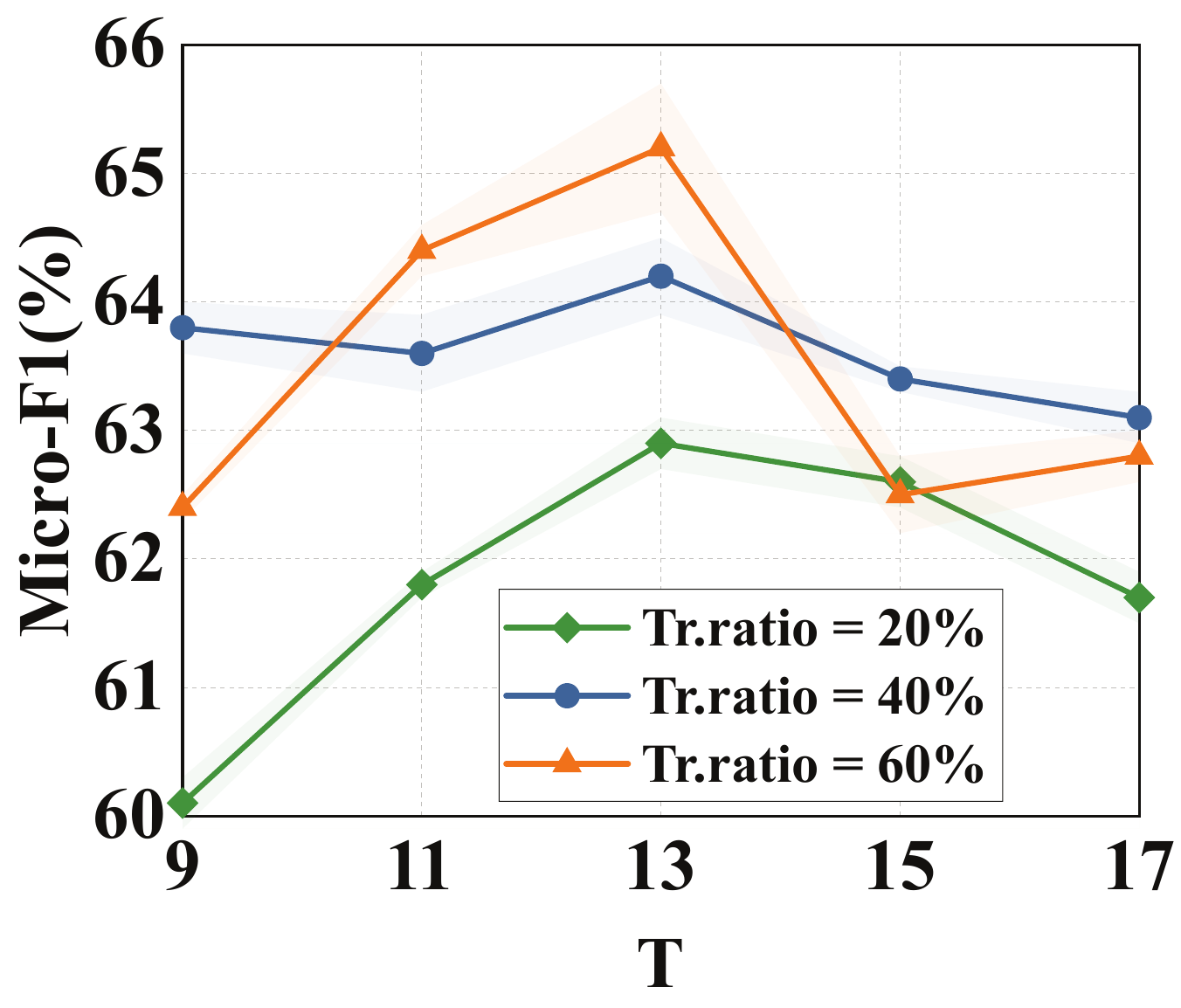}
\end{minipage}%
}%
\centering
\caption{The impact of time steps $T$}
\label{fig6}
\end{figure}

\subsection{Sensitivity Analysis (RQ3)}
This section further studies how different spike neurons and time steps $T$ affect the performance of SpikingHAN, and the experimental results are shown in Fig.~\ref{fig5} and~\ref{fig6}. From the experimental results, we conclude the following: 
\begin{itemize}
    \item \textbf{The impact of spiking neurons.} We use IF \cite{bib31}, LIF \cite{bib32}, and PLIF \cite{bib33} to construct spiking neurons for SNNs, and conduct experiments with a training rate of 20\% on different datasets, the results of which are shown in Fig.~\ref{fig5}. The results indicate that the simple IF neuron is already capable of enabling SpikingHAN to achieve good performance. The LIF neuron improves the performance by adding biologically reasonable leakage terms to IF neuron. The PLIF neuron sets the leakage term in the LIF neuron as a learnable parameter, which confers better flexibility and biological plausibility to the SNNs. Therefore, PLIF performs slightly better than LIF in most cases. 
    \item \textbf{The impact of time steps $T$.} Fig.~\ref{fig6} illustrates the effect of time steps $T$ on the performance of SpikingHAN on different datasets. In fact, the best $T$ value for each dataset in Fig.~\ref{fig6} is not the global optimum, but a local optimum that balances computing cost and performance. This is because as the time steps $T$ increases, the frequency of spiking neurons performing Integrate, Fire, and Reset operations will also increase, and the performance of SpikingHAN may improve, but the running time and memory consumption will also increase, resulting in higher computational costs. Therefore, selecting time steps $T$ should balance computing cost and model performance, and should be considered in combination with the specific requirements of downstream tasks.
\end{itemize}

\section{Conclusion}
In this paper, we propose a novel Spiking Heterogeneous Graph Attention Network model termed SpikingHAN, which incorporates spiking neural networks—known for their brain-inspired and energy-efficient properties—into heterogeneous graph learning, aiming to reduce computational costs while preserving performance. SpikingHAN aggregates metapath-based neighbor nodes through a single-layer graph convolution with shared parameters, and utilizes the semantic-level attention mechanism to aggregate different meta-path semantics. Finally, SNN is used to simulate the spike firing mechanism of biological neurons, encoding heterogeneous information into spike sequences. The pooled spike sequences are then used for prediction, enabling efficient computing with low energy consumption. Experimental results on three real-world datasets indicate that SpikingHAN achieves performance competitive with the best baseline methods using a binary 1-bit representation. At the same time, compared with other heterogeneous graph neural network methods, SpikingHAN shows efficiency advantages in training speed, model parameters, memory usage, and energy consumption. From the perspective of building environment-friendly machine learning models, our work is promising and is expected to inspire further research on efficient heterogeneous graph learning.

\newpage
\appendix
\section{A\hspace{1em}Algorithm}
 Algorithm~\ref{alg1} presents the pseudo-code of SpikingHAN, illustrating the training process of the model. It begins with meta-path-based neighbor aggregation and semantic-level attention to generate comprehensive node representations. These representations are then processed by a spiking neural network (SNN) module, which simulates the membrane potential update, spike firing, and reset mechanisms to achieve brain-inspired and energy-efficient computation.
 
The computational complexity of the meta-path attention module is approximately $\mathcal{O}(P \cdot |\mathcal{V}| \cdot d)$, where $P$ denotes the number of meta-paths, $|\mathcal{V}|$ is the number of nodes, and $d$ represents the embedding dimension. The SNN module introduces temporal dynamics by simulating spiking behavior over $T$ discrete time steps, resulting in an additional complexity of $\mathcal{O}(T \cdot d^2)$. Nevertheless, SpikingHAN achieves high computational efficiency by leveraging binary spike representations and eliminating the need for costly continuous activation functions.
\begin{algorithm}[h]
\caption{Model Training of SpikingHAN}
\label{alg1}
\textbf{Input:} \\ \small
The Heterogeneous Graph $ \mathcal{G}=(\mathcal{V},\ \ \mathcal{E})$ \\
The Initial Features $h^0\in\mathbb{R}^{d_{in}}$ \\
The One-hot matrix of label $y\in\mathcal{V}_L$\\
The meta-path set $\{\mathrm{\Phi}_1,\mathrm{\Phi}_2,...,\mathrm{\Phi}_P\}$\\
\textbf{Output:} \\
Trained model parameters $\theta$ \\
\textbf{Server Executes:} 
\begin{algorithmic}[1] % 数字1表示代码行号将被显示
\State Initialize all parameters in $\theta$
\For{$\mathrm{\Phi}_p\in\{\mathrm{\Phi}_1,\mathrm{\Phi}_2,...,\mathrm{\Phi}_P\}$}
\For{$ i\in\mathcal{V}$}
\State Find the metapath-based neighbors $N_i^{\mathrm{\Phi}_p}$
\State $h_i^{\mathrm{\Phi}_p} \leftarrow$ Aggregate the metapath-based neighbors\\
\hspace*{6.3em}with Eq. (6)
\EndFor
\EndFor
\State Get $\{h^{\mathrm{\Phi}_1},h^{\mathrm{\Phi}_2},...,h^{\mathrm{\Phi}_P}\}$
\State Calculate the importance of each meta-path $I_{\mathrm{\Phi}_p}$ with Eq. (7)
\State The weight of meta-path $\beta_{\mathrm{\Phi}_p} \gets Softmax(I_{\mathrm{\Phi}_p})$
\State $H\gets  \sum_{p=1}^{P}{\beta_{\mathrm{\Phi}_p}\cdot}h^{\mathrm{\Phi}_p}$
\For{$t=1,\ 2,\ ...,T$}
\State Integrate: $V^t\gets  V^{t-1}+\frac{1}{\tau_m}\left(dropout(H\cdot W_3)\right.$\\
\hspace*{5.5em}$\left.-(V^{t-1}-V_{th})\right)$
\State Fire: $\mathrm{\Theta}\left(V^t\right)\gets$  1 if $V^t\geq V_{th}$ else 0
\State Reset: $V^t\gets  \mathrm{\Theta}\left(V^t\right)\cdot(V^t-V_{th})+(1-\mathrm{\Theta}\left(V^t\right))\cdot V^t$
\EndFor
\State $\hat{y}\gets  \frac{1}{\left|T\right|}\sum_{t=1}^{T}\mathrm{\Theta}\left(V^t\right)$
\State Loss $\mathcal{L}\gets  -\sum_{i\in\mathcal{V}_L} y_i\cdot ln\left({\hat{y}}_i\right)$
\State Model optimization, update the parameters $\theta$
\State \Return{trained model with parameters $\theta$}
\end{algorithmic}
\end{algorithm}

\section{B\hspace{1em}Dataset Details}
The experiments employ three commonly used heterogeneous graph datasets to evaluate the performance of SpikingHAN. The detailed characteristics of the datasets are summarized in Table~\ref{tab1}. 
\begin{itemize}
    \item DBLP is an English literature dataset in computer science. After data preprocessing, a subset was extracted, comprising 14,328 papers, 4,057 authors, 20 venues, and 7,723 terms. The authors are categorized into four research fields, including Databases, Data Mining, Information Retrieval, and Artificial Intelligence. The initial representation of each author is obtained by bag-of-word encoding of the keywords of their paper. In addition, we perform experiments based on the predefined meta-paths set \{APA, APVPA, APTPA\}.
    \item ACM is a literature dataset covering various subjects in computer science. The dataset is sourced from \cite{bib10} and contains 3,025 papers, 5,825 authors, and 26 subjects. The papers are categorized into three classes, including Wireless Communications, Databases, and Data Mining. The initial representation of each paper is obtained by bag-of-word encoding of its keywords. In addition, we perform experiments based on the predefined meta-paths set \{PAP, PSP\}.
    \item IMDB is a dataset about TV shows, movies, and related people information. A subset was extracted and preprocessed, resulting in a dataset containing 4,278 movies, 2,081 directors, and 5,257 actors. The movies are categorized into three classes, including Action, Comedy, and Drama. The initial representation of each movie is obtained by bag-of-words encoding of its plot keywords. In addition, we perform experiments based on the predefined meta-paths set \{MDM, MAM\}.
\end{itemize}

\begin{table}[t]
\centering
\small
\renewcommand{\arraystretch}{1.3}
\begin{tabular}{c|l|c|c}
\hline
\textbf{Datasets} & \multicolumn{1}{c|}{\textbf{Nodes}} & \multicolumn{1}{c|}{\textbf{Edges}} & \textbf{Meta-paths} \\ \hline
\multirow[c]{4}{*}{DBLP} 
 & Paper (P): 14,328        & A-P: 19,645          & APA        \\
 & Author (A): 4,057        & P-V: 14,328          & APVPA      \\
 & Venue (V): 20            & P-T: 85,810          & APTPA      \\
 & Term (T): 7,723          &                      &            \\ \hline
\multirow[c]{3}{*}{ACM}  
 & Paper (P): 3,025         & P-A: 9,744           & PAP        \\
 & Author (A): 5,825        & P-S: 3,025           & PSP        \\
 & Subject (S): 26          &                      &            \\ \hline
\multirow[c]{3}{*}{IMDB} 
 & Movie (M): 4,278         & M-D: 4,278           & MDM        \\
 & Director (D): 2,081      & M-A: 12,828          & MAM        \\
 & Actor (A): 5,257         &                      &            \\ \hline
\end{tabular}
\caption{The description of datasets}
\label{tab1}
\end{table}

\section{C\hspace{1em}Baselines and experimental setup}
The details of baselines are summarized as follows:
\begin{itemize}
    \item \textbf{GAT} \cite{bib36}: The model introduces an attention mechanism in the graph neighbor node aggregation operation, assigning weights to neighboring nodes and combining their feature information in a weighted manner.
    \item \textbf{DAGNN} \cite{bib37}: The model separates transformations and message propagation and balances local neighborhood information with global neighborhood information through an adaptive regulation mechanism.
    \item \textbf{SpikingGCN} \cite{bib15}: The model combines graph convolution with spiking neural networks, effectively merging convolutional features into spiking neurons.
    \item \textbf{SpikeGCL} \cite{bib16}: The model applies SNNs to graph comparison learning, which enables binarized representation learning on graphs via SNNs.
    \item \textbf{HAN} \cite{bib10}: The model utilizes two-layer attention mechanisms to process heterogeneous graphs, effectively generates node embeddings by hierarchically aggregating neighbor features.
    \item \textbf{HINormer} \cite{bib38}: The model captures diverse information in heterogeneous graphs through local structure encoders and heterogeneous relationship encoders, thereby achieving comprehensive node representation.
    \item \textbf{PHGT} \cite{bib21}: The model employs a novel multi-token design, including node, semantic, and global tokens, to effectively capture higher-order heterogeneous semantic relationships and long-range dependencies in heterogeneous graphs.
\end{itemize}

The SpikingHAN model is optimized using the Adam optimizer. The number of training epochs is configured to 200, with an early stopping mechanism that has a patience value of 100. The SNN part uses PLIF neurons and resets the spike neurons by subtracting the threshold. The classification evaluation metrics adopt Micro-F1 and Macro-F1, combining these two metrics can more comprehensively evaluate the classification capability of the model, considering both the overall performance and the balance of each class. GAT, DAGNN, SpikingGCN, and SpikeGCL are applied on metapath-based homogeneous graphs and the classification results under the best meta-path are reported. In the three datasets, the classification nodes are split into training, validation, and testing sets based on three ratios which are (20\%, 10\%, 70\%), (40\%, 10\%, 50\%), and (60\%, 10\%, 30\%). Both SpikingHAN and baseline methods use the same training, validation, and test sets and are trained with 10 different random seeds, and finally report the mean and standard deviation.

\section{Acknowledgments}
The work is supported by the National Natural Science Foundation of China (No. 62572186) and the Open Project Funding of the Key Laboratory of Intelligent Sensing System and Security (Hubei University), Ministry of Education.

\bibliography{aaai2026}

\end{document}